# Vehicle-in-Virtual-Environment Method for ADAS and Connected and Automated Driving Function Development/Demonstration/Evaluation

Xincheng Cao, Haochong Chen, Bilin Aksun-Guvenc, Levent Guvenc

Automated Driving Lab, Ohio State University

## Abstract

The current approach for new Advanced Driver Assistance System (ADAS) and Connected and Automated Driving (CAD) function development involves a significant amount of public road testing which is inefficient due to the number miles that need to be driven for rare and extreme events to take place, thereby being very costly also, and unsafe as the rest of the road users become involuntary test subjects. A new development, evaluation and demonstration method for safe, efficient, and repeatable development, demonstration and evaluation of ADAS and CAD functions called Vehicle-in-Virtual – Environment (VVE) was recently introduced as a solution to this problem. The vehicle is operated in a large, empty, and flat area during VVE while its localization and perception sensor data is fed from the virtual environment with other traffic and rare and extreme events being generated as needed. The virtual environment can be easily configured and modified to construct different testing scenarios on demand. This paper focuses on the VVE approach and introduces the coordinate transformations needed to sync pose (location and orientation) in the virtual and physical worlds and handling of localization and perception sensor data using the highly realistic 3D simulation model of a recent autonomous shuttle deployment site in Columbus, Ohio as the virtual world. As a further example that uses multiple actors, the use of VVE for Vehicle-to-VRU communication based Vulnerable Road User (VRU) safety is presented in the paper using VVE experiments and real pedestrian(s) in a safe and repeatable manner. VVE experiments are used to demonstrate the efficacy of the method.

## Introduction

In recent years, most modern cars have been equipped with Advanced Driver Assistance Systems (ADAS) **[1]**. These systems not only enhance driving safety and convenience but also are a significant step leading towards fully autonomous driving which will help solve the end mile problem in smart cities **[2]**. As a result, there has been plenty of research conducted on developing, improving and evaluating ADAS and active safety in references **[3]**–**[6]**. Additionally, **[7]** provides a comprehensive explanation of the details regarding to develop ADAS which offer a solid theoretical foundation provided for following research in this field.

In general, current ADAS system development procedure utilized by OEM includes the following steps in sequence: 1) extensive model-in-the-loop (MIL) simulation; 2) hardware-in-the-loop (HIL) testing; 3) proving ground testing; 4) public road testing. As the testing proceeds, more hardware and realism are gradually introduced into the system. For example, model-in-the-loop (MIL) is a purely virtual simulation tool which is introduced in detail in **[8]**, while hardware-in-the-loop (HIL) starts to incorporate hardware components into the setup which is discussed in detail in **[9], [10]**. Both MIL and HIL testing require both low and high fidelity, validated vehicle models that incorporate longitudinal **[11]**, lateral **[12]**, and vertical dynamics **[13]**. This is not needed in the VVE approach as the actual vehicle is used.

It is evident that the public road-testing part described above is very expensive and cost ineffective as encountering rare and extreme cases may require millions of miles of driving, not to mention that public road testing automatically involves other traffic users as involuntary participants of the test, which may cause safety concerns. As a result, this paper introduces vehicle-in-virtual-environment (VVE) as an intermediate approach to the procedure listed above, to be employed before public road testing. The proposed method combines the advantages of realism resulting from real vehicle dynamics and overall testing safety. Additionally, the VVE method can allow the safe testing of rare traffic scenarios and cases without requiring long periods of running on public road as these can easily be injected into the virtual environment.

One research and application area that can take particular advantages of the VVE method is vehicle-to-pedestrian (V2P) communication for Vulnerable Road User (VRU) safety. V2P communication uses the exchange of information between vehicles and pedestrians to ensure traffic safety. Through communication protocols like Dedicated Short Range Communications (DSRC) based on IEEE 802.11p standard discussed in **[14]**, cellular networks based on 3GPP standards reviewed in **[15]**, C-V2X based on 4G/5G/6G or Bluetooth communicated information can be exchanged between the vehicles and mobile devices carried by pedestrians such as cellphones. There have been several implementation cases for V2P safety, such as **[16]** that uses a mobile phone application to broadcast personal safety messages (PSM) via Bluetooth communication.

The outline of the rest of the paper is as follows. Following related autonomous testing methodologies and a concise overview of VVE in the Introduction, the VVE System Architecture section elaborates on VVE system and hardware architecture. The next section provides an extensive discourse on the VVE algorithm with particular focus on frame transformations and the configuration of the software environment. The subsequent section demonstrates the application scenario of VVE to test V2P, illustrating the functionality of the VVE method in evaluation of AV testing for VRU safety. This is followed by a dive into the architecture of Vehicle-to-Pedestrian (V2P) communication, examining how this framework can be strategically employed to assess pedestrian safety. The next section demonstrates the outcomes of various tests of V2P using VVE method, accompanied by a comprehensive discussion analyzing these results. The paper culminates with conclusions in the last section where recommendations for future work are also presented.

## VVE System Structure

The architecture of current autonomous vehicles used for public road testing is displayed in Figure 1. Onboard sensors on the AV collect information from the environment, and data-processing routines such as perception and situational awareness algorithms are then run to generate data for the decision-making module, which subsequently generates high-level control commands such as path generation/modification and acceleration profile. A control unit is then used to generate low-level control commands (such as throttle, brake and steering commands) that are transmitted to the Drive-by-Wire equipped AV via CAN connection. It should be noted that this chain of operations takes place purely in the real world and is how AVs operates in the real world.

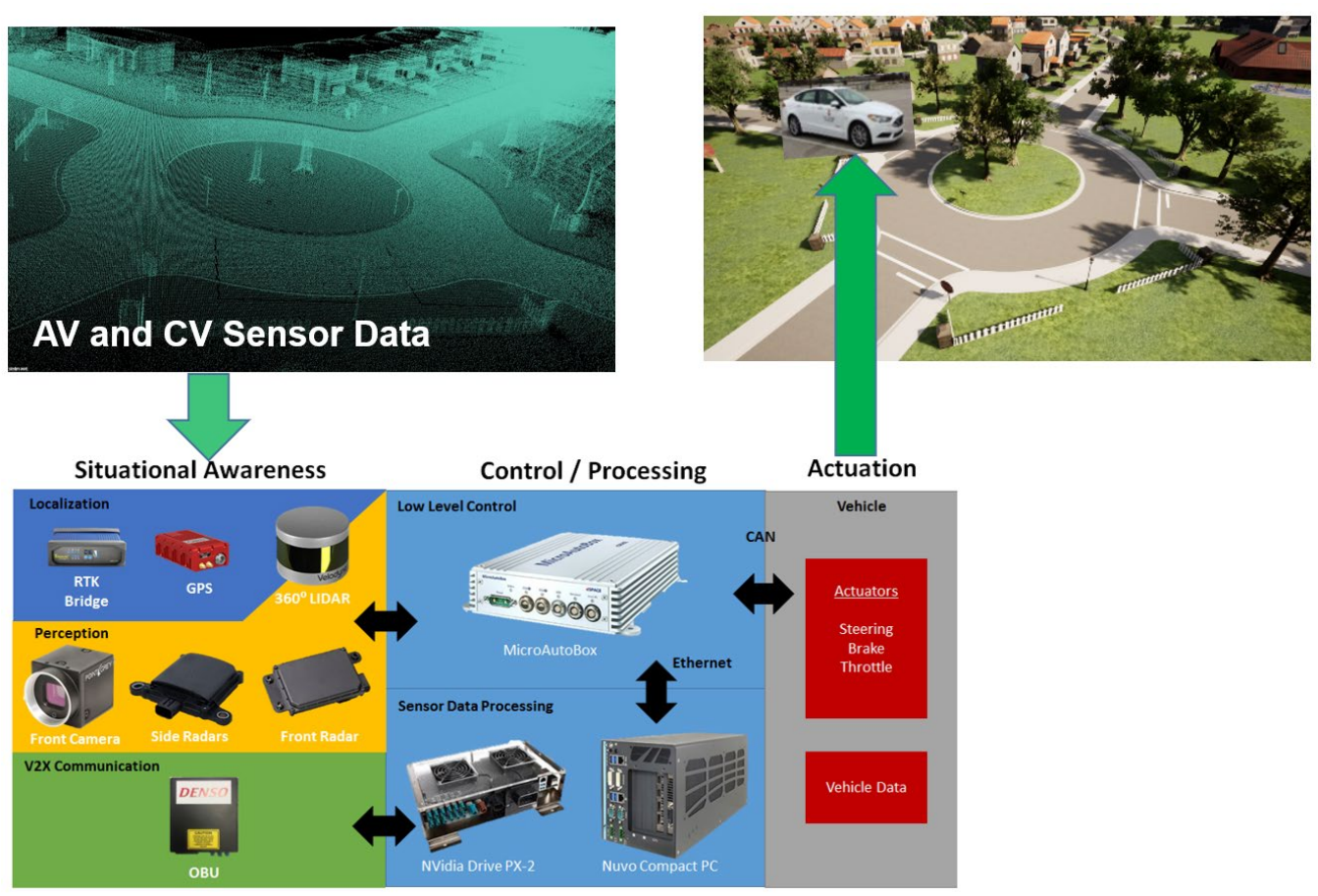

Figure 1. Current AV Architecture [17]

The VVE architecture is illustrated in Figure 2. The real vehicle operates in a safe and empty space such as a parking lot, and its motions are synchronized to those of a virtual vehicle located in a highly realistic 3D-rendered virtual environment that can be easily modified. The real sensors onboard the real vehicle are also replaced by virtual sensor feeds that are generated in the virtual environment from the perspective of the motion-synchronized virtual vehicle. As such, the control unit onboard the real vehicle will react to the traffic scenarios represented in the virtual environment and deliver corresponding control commands to the real vehicle, and its motions are in turn synchronized into the virtual vehicle, completing the loop.

Comparing the traditional public road-testing approach with the proposed VVE method, several observations can be reached. For the public road-testing, other road users are involuntarily included in the testing process, as the test vehicle operates in public traffic, which poses as a potential safety concern. Also, the occurrence of specific traffic scenarios depends on chance, as specific scenarios of interest cannot be generated on demand during public road driving. This also means that rare-occurring, difficult and potentially dangerous scenarios cannot be comprehensively tested without extensive driving mileage, which not only poses safety concerns due to the involvement of real traffic users but also is neither time nor cost efficient. Comparatively, the traffic scenarios in the virtual environment can be easily modified for the VVE approach. This means that safety critical boundary cases can easily be replicated and tested in a safe manner, as the real vehicle operates in a safe open space. Also due to the ease of creation of various traffic cases, the costs to test the AV systems using this approach are expected to be much lower, as the testing procedure can be made much more efficient and less time-consuming. An additional benefit of the VVE method is that it utilizes the real vehicle and hence its physical dynamics instead of depending on theoretical vehicle models used in HIL/MIL testing methods, making it more realistic and reliable. Furthermore, if a real-world example of the vehicle of interest is not available for testing, the available test vehicle can be made to emulate the dynamic behavior of the desired type of vehicle. An extra comment on the VVE method is that it aligns well with current regulatory testing and also offers more possibilities, as many tests, such as the 'sine with dwell' test for yaw stability validation, can easily be prepared and implemented using VVE. VVE can also be used to test and evaluate AV operations in geo-fenced areas before granting permission to operate.

It is worth noting, though, that apart from the above-mentioned advantages of the VVE method, certain limitations do still exist for this approach. For example, since the real vehicle is operated in a physical empty space, the virtual environment needs to be divided into blocks, with each virtual block being no larger than the real testing space available. As a result, in order to traverse from one virtual block to another, virtual roundabouts must be added to connect neighboring blocks so that the real vehicle has the opportunity to turn around when it reaches the edge of physical testing space.

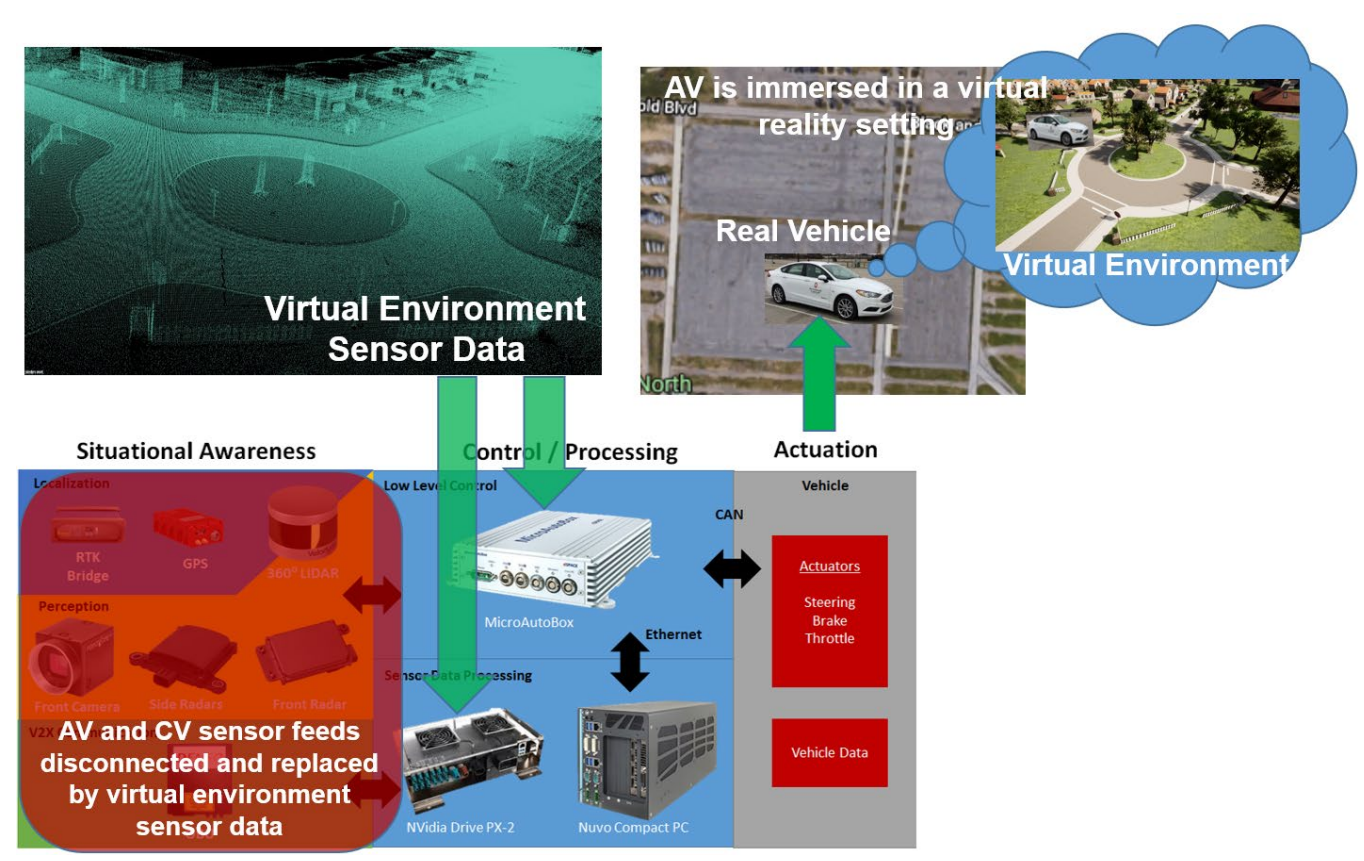

Figure 2. VVE Architecture [17]

Our current implementation architecture of VVE is illustrated in Figure 3, and the components onboard our test vehicle are displayed in Figure 4. The real vehicle is equipped with an RTK GPS unit (OxTS xNAV500) with differential antennas that provide both position and heading information even when the vehicle is stationary. These data are received and collected by the onboard control unit, dSpace microautobox (MABX), and transmitted to the onboard simulation computer that runs an Unreal based (CARLA) virtual environment via Ethernet UDP Protocol. The virtual environment used in our implementation is the replication of the Linden area, where a recent autonomous shuttle deployment took place in Columbus, Ohio **[18]**. The virtual environment receives the real GPS data and applies frame transformation/conversion operations to synchronize the virtual vehicle motions with those of the real vehicle. The virtual sensors collect measurements in the virtual environment and transmit the information back to the MABX via Ethernet UDP Protocol. It should be noted that virtual perception sensor data, such as virtual radar and lidar data, will be transmitted directly back to MABX, while virtual localization sensor data, such as virtual GPS positions, require a set of inverse frame transformation/conversion operations before being sent back to MABX. The MABX can then generate actuator signals from the received information and control the real vehicle via CAN bus connection. It should be noted that real-virtual frame transformation/conversion operations are critical to the satisfactory motion synchronization of the real and virtual vehicle, and hence will be discussed in detail in a separate section.

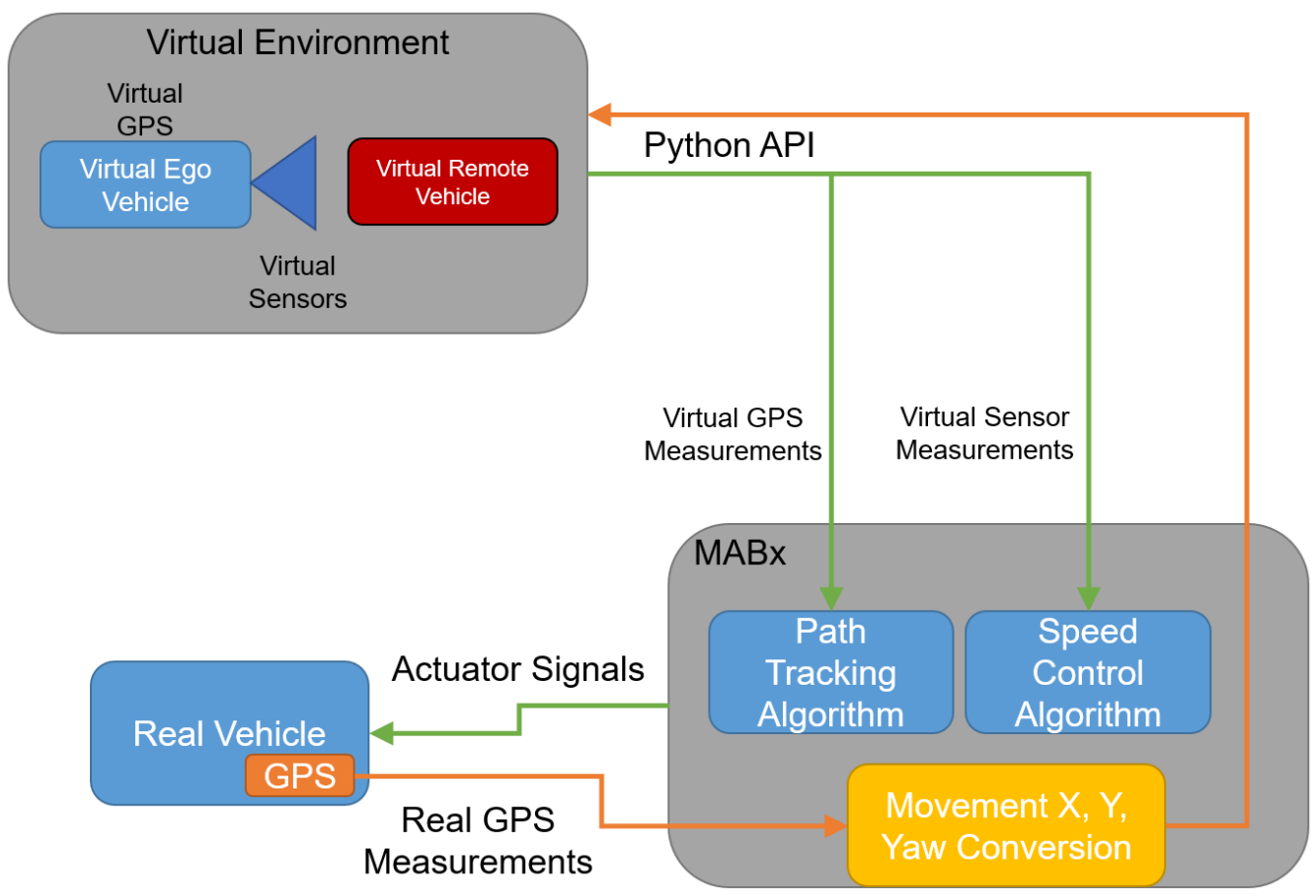

Figure 3. Implemented VVE architecture. **[17]**

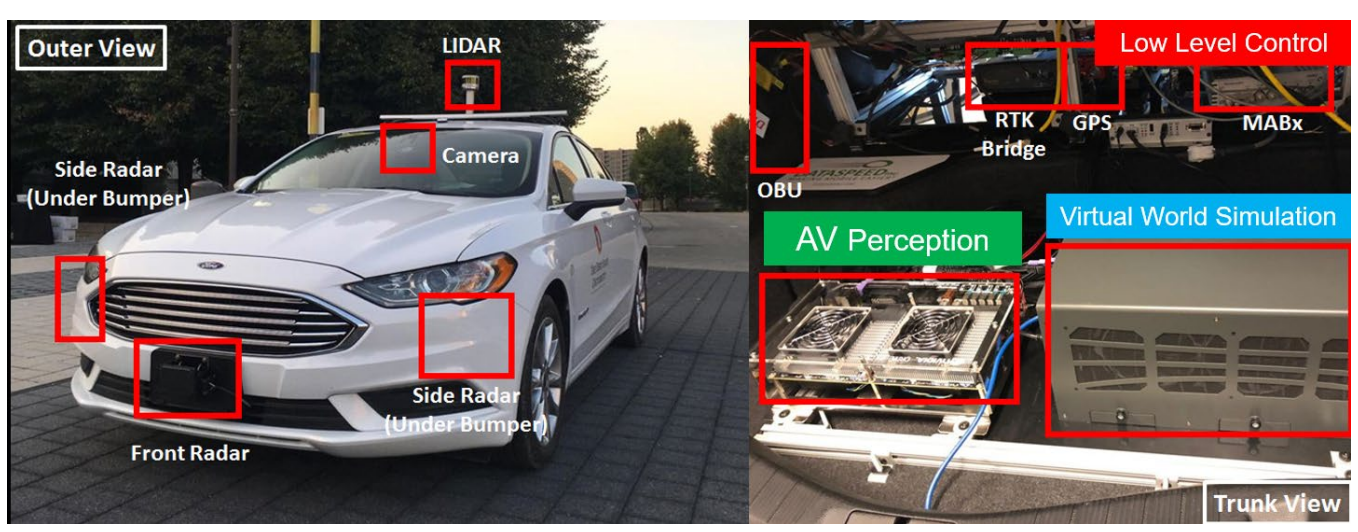

Figure 4. AV used for VVE implementation **[17]**

## Frame Transformation/Conversion for Motion Synchronization

As mentioned in the above section, frame transformation and conversion operations are crucial to the accurate motion synchronization of the real and virtual vehicles. Hence, this section aims to demonstrate the procedure we employed to represent the real-world motions in the virtual environment.

In general, motion synchronization begins at a certain time, defined as ‘reset time’, where the virtual vehicle location and heading are initialized to a predetermined set of values, while the real vehicle location and heading at this time are recorded for later use. As the real vehicle moves, the changes in its locations and heading compared to those at reset time undergo a set of frame transformation/conversion operations so that these changes represented in the appropriate virtual frame can be applied in the virtual environment, hence achieving motion synchronization. It should be noted that once the vehicle center-of-gravity position and heading is resolved with frame transformation/conversion process, the coordinates of the entire vehicle body can be reconstructed accordingly as well. The frame transformation/conversion process consists of multiple components, and each of them will be described separately in the following paragraphs.

The first component of the process is a frame conversion operation illustrated in Figure 5. The OxTS GPS provides the longitude, latitude and heading angle data in degrees, where the heading angle is recorded in the range of -180 deg and 180 deg. Converting the longitude and latitude data into unit of meter, one can construct a coordinate frame, namely the real vehicle (Fr) frame, as displayed in the left-hand side of Figure 5. It should be noted that the positive X-axis in this frame is assumed to be aligned with positive latitude direction, where zero heading angle corresponds to. It should also be noted that the positive heading angle in this frame is defined as clockwise (CW) in correspondence to the heading angle outputted by the OxTS GPS unit. In order to accommodate frame transformation operations in other components of the process, a new frame named GPS (Fg) frame is constructed as displayed in the right-hand side of Figure 5. In this Fg frame, zero heading is defined to be along the positive X-axis and positive heading angle is in the counterclockwise (CCW) direction. The frame conversion can thus be carried out as follows:

$$\begin{cases} \boldsymbol{X_g} = \boldsymbol{X_r} \\ \boldsymbol{Y_g} = -\boldsymbol{Y_r} \\ \boldsymbol{\psi_g} = -\boldsymbol{mod}(\boldsymbol{\psi_r} + \mathbf{360}, \mathbf{360}) \end{cases} \quad (1)$$

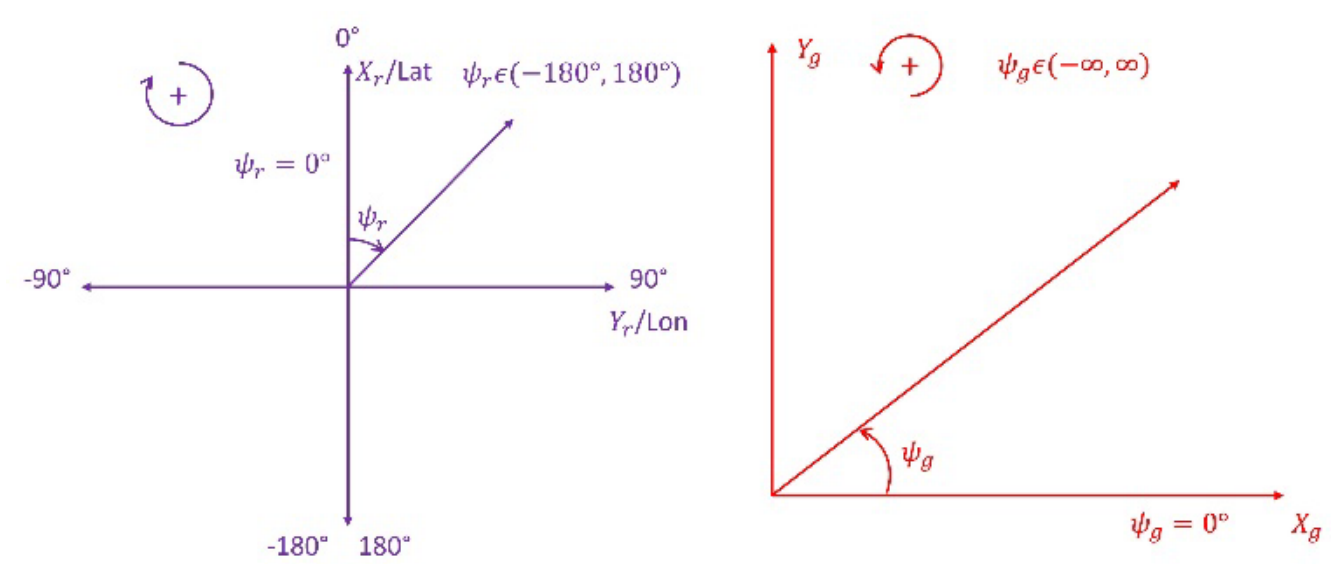

Figure 5. Frame transformation/conversion component 1: conversion between Fr frame and Fg frame

The second component of the process is a frame transformation operation illustrated in Figure 6. The Fg frame is the same as described in Figure 5. At reset time t0, an intermediate (F) frame is created at the vehicle location with positive X-axis pointing at zero heading direction and positive heading defined in the CCW direction. As the vehicle moves to a new location at time t1, its location and heading can be represented in the F frame as illustrated in Equation 2. It should be noted that set $(\boldsymbol{X_{g0}}, \boldsymbol{Y_{g0}}, \boldsymbol{\psi_{g0}})$ and set $(\boldsymbol{X_{g1}}, \boldsymbol{Y_{g1}}, \boldsymbol{\psi_{g1}})$ can be obtained by using Equation 1 on vehicle states at the reset time and at current time, respectively.

$$\begin{cases} \boldsymbol{\psi_1 = \psi_0 + \psi_{g1} - \psi_{g0}} \\ \begin{bmatrix} \boldsymbol{X_1} \\ \boldsymbol{Y_1} \end{bmatrix} = \begin{bmatrix} \boldsymbol{X_0} \\ \boldsymbol{Y_0} \end{bmatrix} + \boldsymbol{R_{Fg\to F}} \begin{bmatrix} \boldsymbol{X_{g1} - X_{g0}} \\ \boldsymbol{Y_{g1} - Y_{g0}} \end{bmatrix} \end{cases} \quad (2)$$

$$\text{Where: } \boldsymbol{R_{Fg\to F}} = \begin{bmatrix} \boldsymbol{cos\,(\psi_{g0})} & \boldsymbol{sin(\psi_{g0})} \\ \boldsymbol{-sin\,(\psi_{g0})} & \boldsymbol{cos(\psi_{g0})} \end{bmatrix}$$

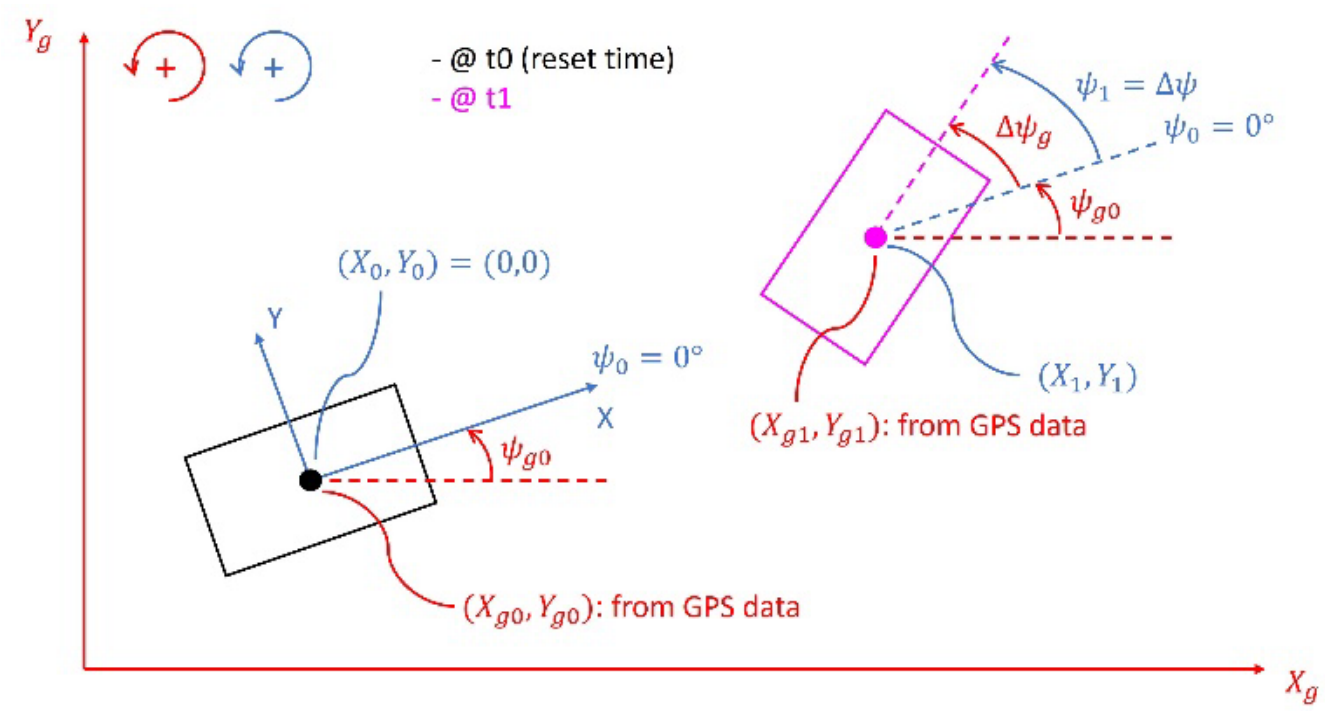

Figure 6. Frame transformation/conversion component 2: transformation between Fg frame and F frame

The third component of the process is another frame conversion operation displayed in Figure 7. The coordinate frame used in the virtual environment, named as the Fc frame, is shown on the right-hand side of Figure 7, where the initial location and heading of the virtual vehicle is defined at reset time t0. An additional coordinate frame, named the vehicle (Fv) frame, is defined as shown on the left-hand side of Figure 7, with zero heading pointing in the positive X-axis direction and positive heading angle in the CCW direction. The frame transformation operation can thus be carried out as follows:

$$\begin{cases} \boldsymbol{X_c = X_v} \\ \boldsymbol{Y_c = -Y_v} \\ \boldsymbol{\psi_c = -\psi_v} \end{cases} \quad (3)$$

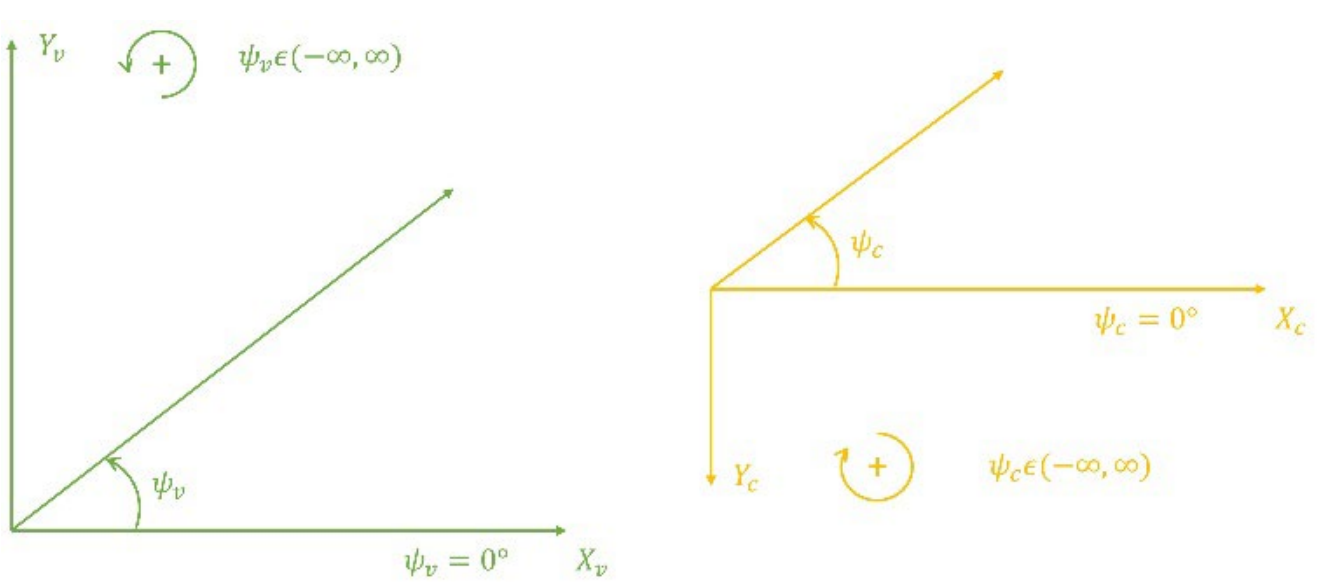

Figure 7. Frame transformation/conversion component 3: conversion between Fv frame and Fc frame

The final component of the process is illustrated in Figure 8. This frame transformation procedure connects the previously discussed intermediate F frame and vehicle Fv frame. The location and heading of the vehicle at time t1 can be represented in the Fv frame as illustrated in Equation 4. It should be noted that the set $(\boldsymbol{X_{v0}, Y_{v0}, \psi_{v0}})$ can be obtained by using Equation 3 on the initial position and heading of the virtual vehicle defined in the Fc frame. Once $(\boldsymbol{X_{v1}, Y_{v1}, \psi_{v1}})$ set is obtained from Equation 4, Equation 3 can be applied again to represent the current vehicle location and heading information in the Fc frame, which will complete the motion synchronization procedure.

$$\begin{cases} \boldsymbol{\psi_{v1} = \psi_{v0} + \psi_1 - \psi_0} \\ \begin{bmatrix} \boldsymbol{X_{v1}} \\ \boldsymbol{Y_{v1}} \end{bmatrix} = \begin{bmatrix} \boldsymbol{X_{v0}} \\ \boldsymbol{Y_{v0}} \end{bmatrix} + \boldsymbol{R_{F\to Fv}} \begin{bmatrix} \boldsymbol{X_1 - X_0} \\ \boldsymbol{Y_1 - Y_0} \end{bmatrix} \end{cases} \quad (4)$$

$$\text{Where: } \boldsymbol{R_{F\to Fv}} = \begin{bmatrix} \boldsymbol{cos\,(\psi_{v0})} & \boldsymbol{-sin(\psi_{v0})} \\ \boldsymbol{sin\,(\psi_{v0})} & \boldsymbol{cos(\psi_{v0})} \end{bmatrix}$$

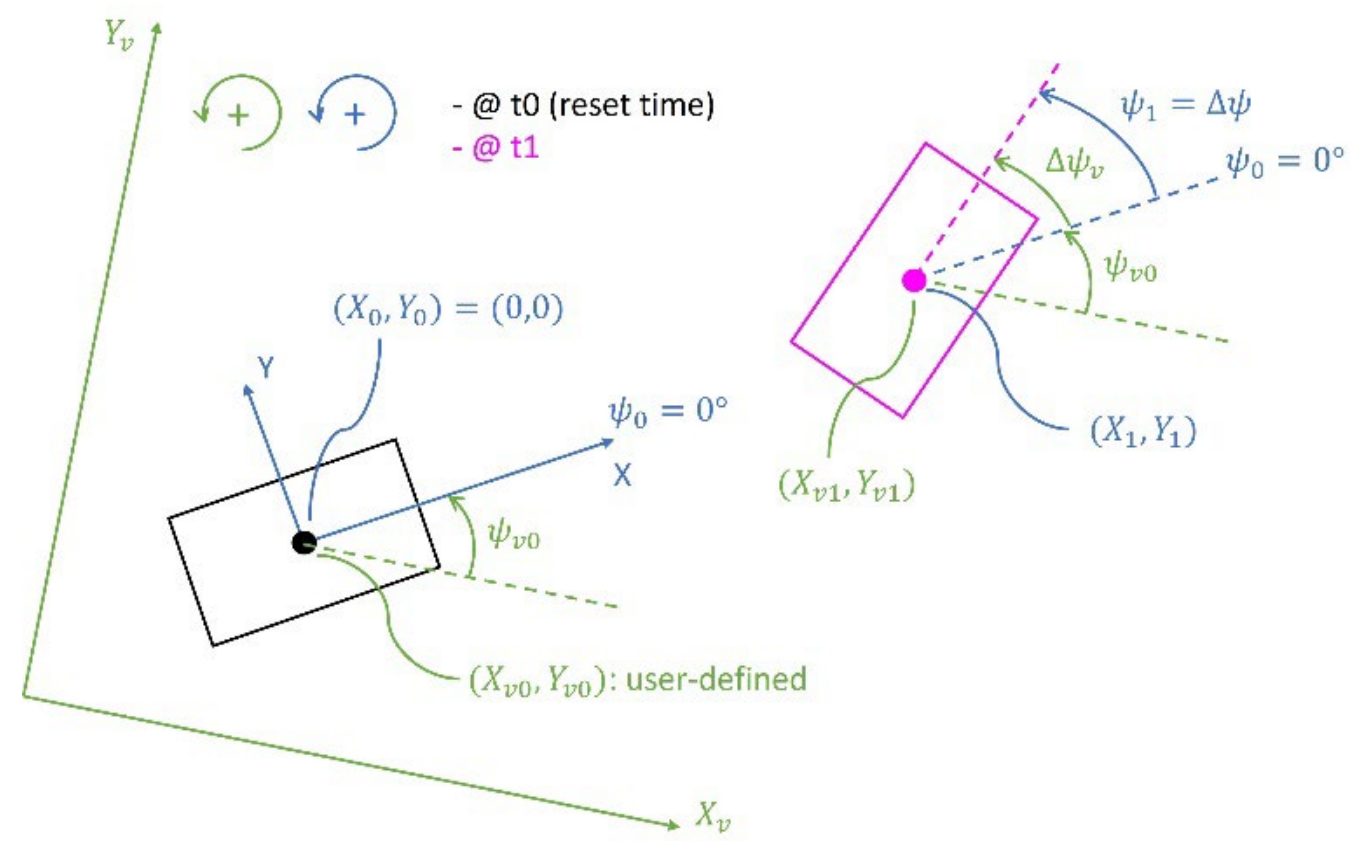

Figure 8. Frame transformation/conversion component 4: transformation between F frame and Fv frame

## V2P System Structure

To illustrate the multi-actor capabilities of the proposed VVE approach, a Vehicle-to-VRU communication is implemented. The system architecture is displayed in Figure 9. The pedestrian carries a mobile phone equipped with onboard IMU and GPS sensors. A mobile application, adapted from [16], is implemented on the phone, where the GPS location and heading of the pedestrian are broadcasted as part of personal safety messages (PSM) via Bluetooth low-energy (BLE) connection. A Bluetooth reception board listens to the PSM and transmits pedestrian motion information to the virtual environment via Ethernet connection between the board and the simulation computer onboard the real vehicle. Frame transformation/conversion operations similar to those described in the previous section are carried out in the virtual environment to synchronize the virtual pedestrian motions to the real pedestrian motions. With this architecture, it is possible to initialize virtual pedestrian position and orientation in the virtual environment such that traffic testing scenarios can be constructed, while the real pedestrian operates in a safe space out of the harm's way, ensuring the safety of the test. Figure 10 demonstrates one of such possible test setups. Figure 11 presents the information flow of our VVE V2P implementation.

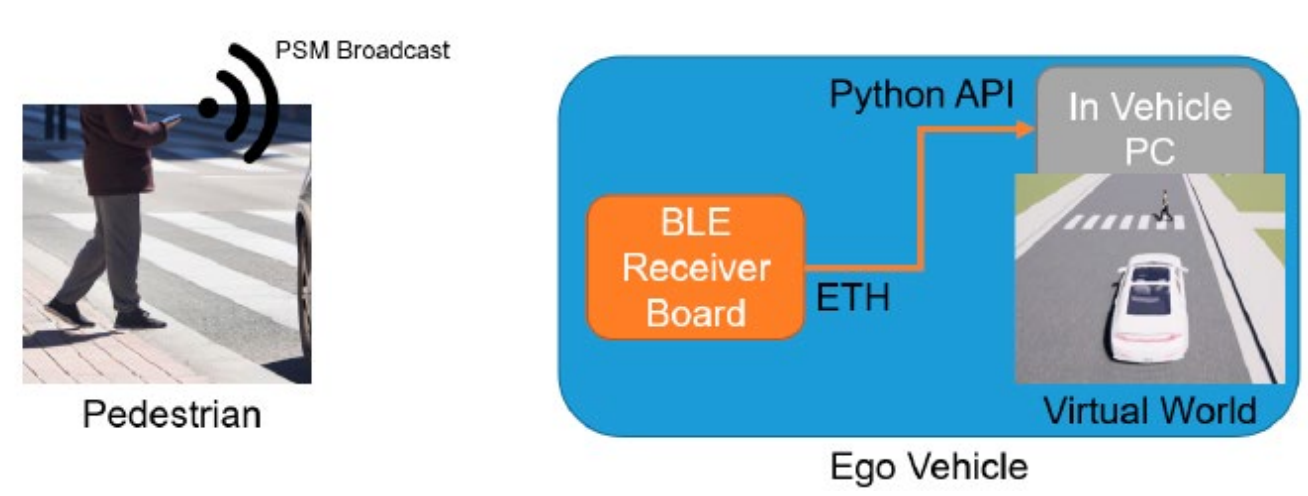

Figure 9. Implemented Vehicle-to-VRU architecture [17]

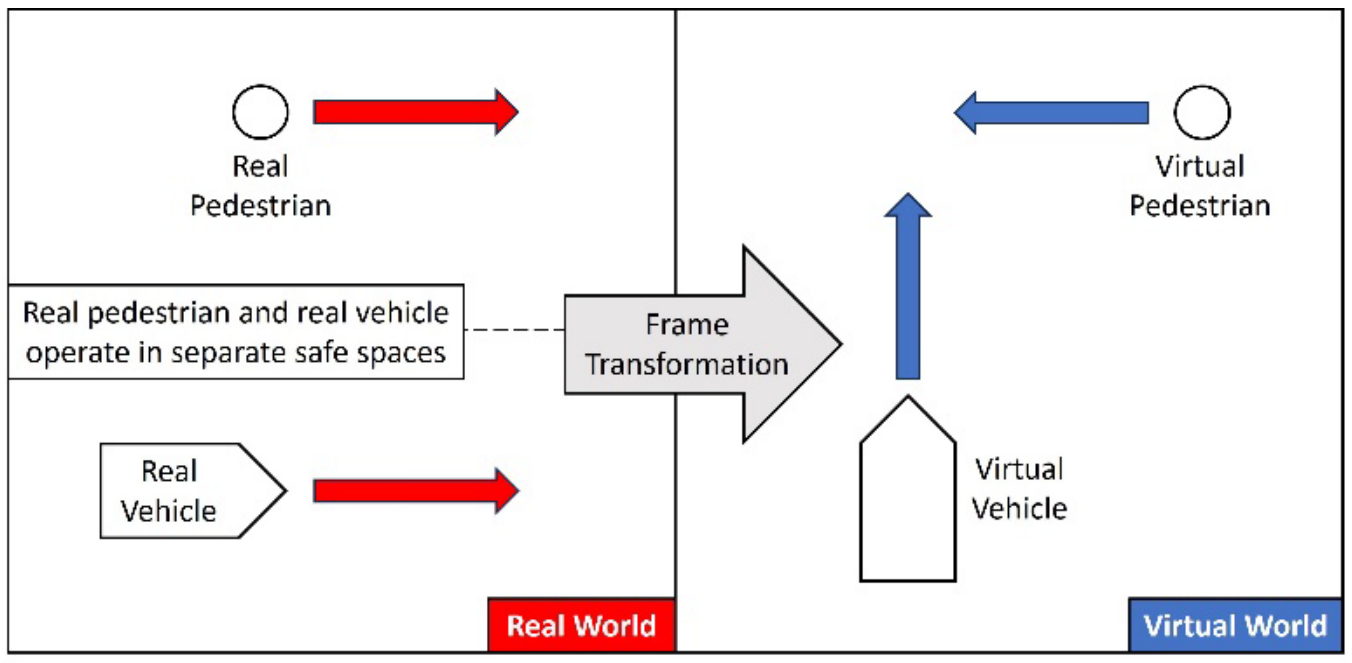

Figure 10. One possible V2P test setup with VVE method. The arrow directions represent the actors' directions of travel in the real (red) and the virtual (blue) world.

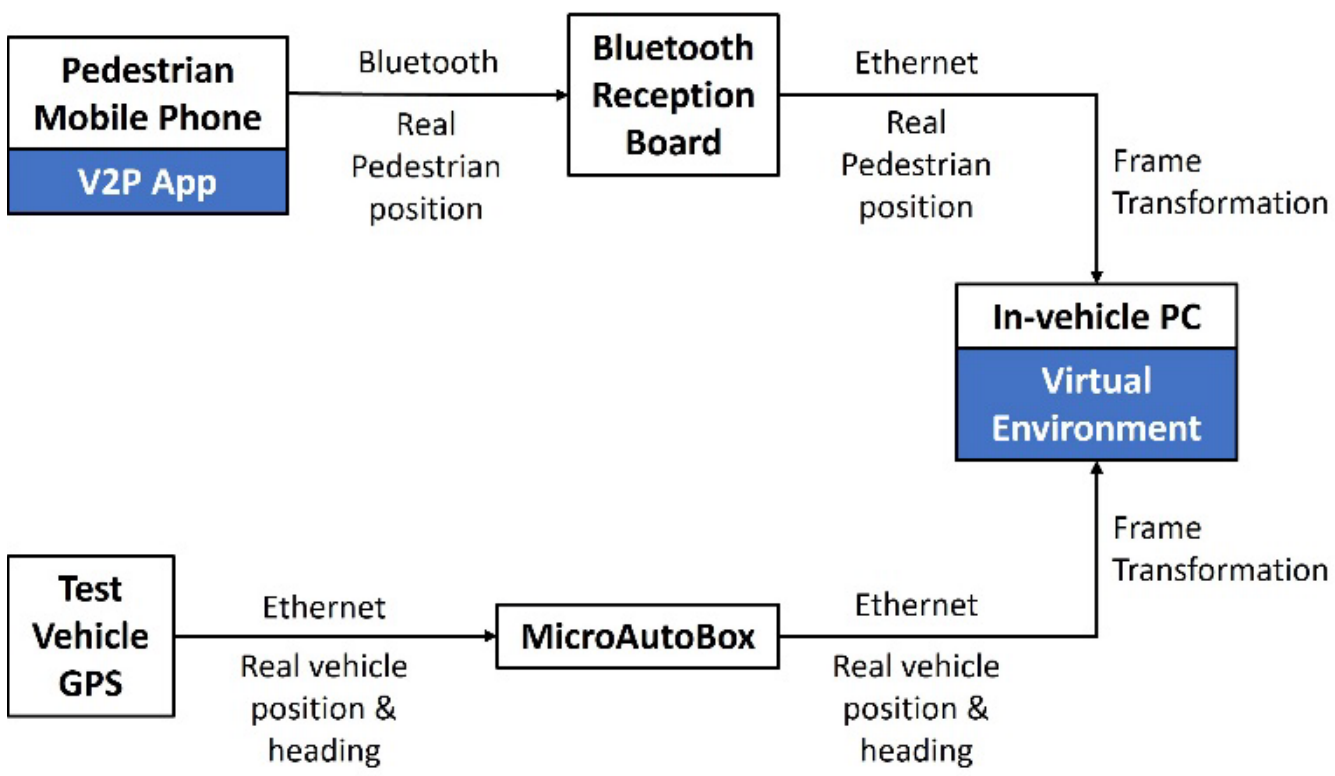

Figure 11. VVE V2P information flow

## Implementation Results of V2P Communication in VVE

The capability of real-virtual vehicle motion synchronization of the proposed VVE method is tested using the setup shown in Figure 12. The screen in the vehicle is connected to the simulation computer and displays the virtual vehicle motion in the virtual environment, while the real test vehicle is operating in an open parking lot. Testing results are displayed in Figure 13, where the motion trajectory of the real vehicle and the motion trajectory of the virtual vehicle are overlayed on the real-world coordinate frame and the virtual environment map, respectively. It can be observed that satisfactory motion synchronization between the real-world motion and virtual world motion has been achieved. Virtual sensor data generated in CARLA environment along the test route is displayed in Figure 14, where onboard RGB camera and LiDAR data are shown (the sensor data videos can be found as follows: https://youtu.be/Sw3W-sQOOqk; https://youtu.be/Pc2PyHEC8E8). It should be noted that these virtual sensor data can be fed back to the real vehicle as perception information so that the real automated vehicle can react to the virtual scenario. Other virtual sensors can also be easily added to the virtual vehicle in the virtual environment. It is worth noting, however, that during the test, the real vehicle is being manually driven in the open parking space at a low speed, since the goal of this test is to safely demonstrate the real-virtual synchronization capability of the proposed VVE method.

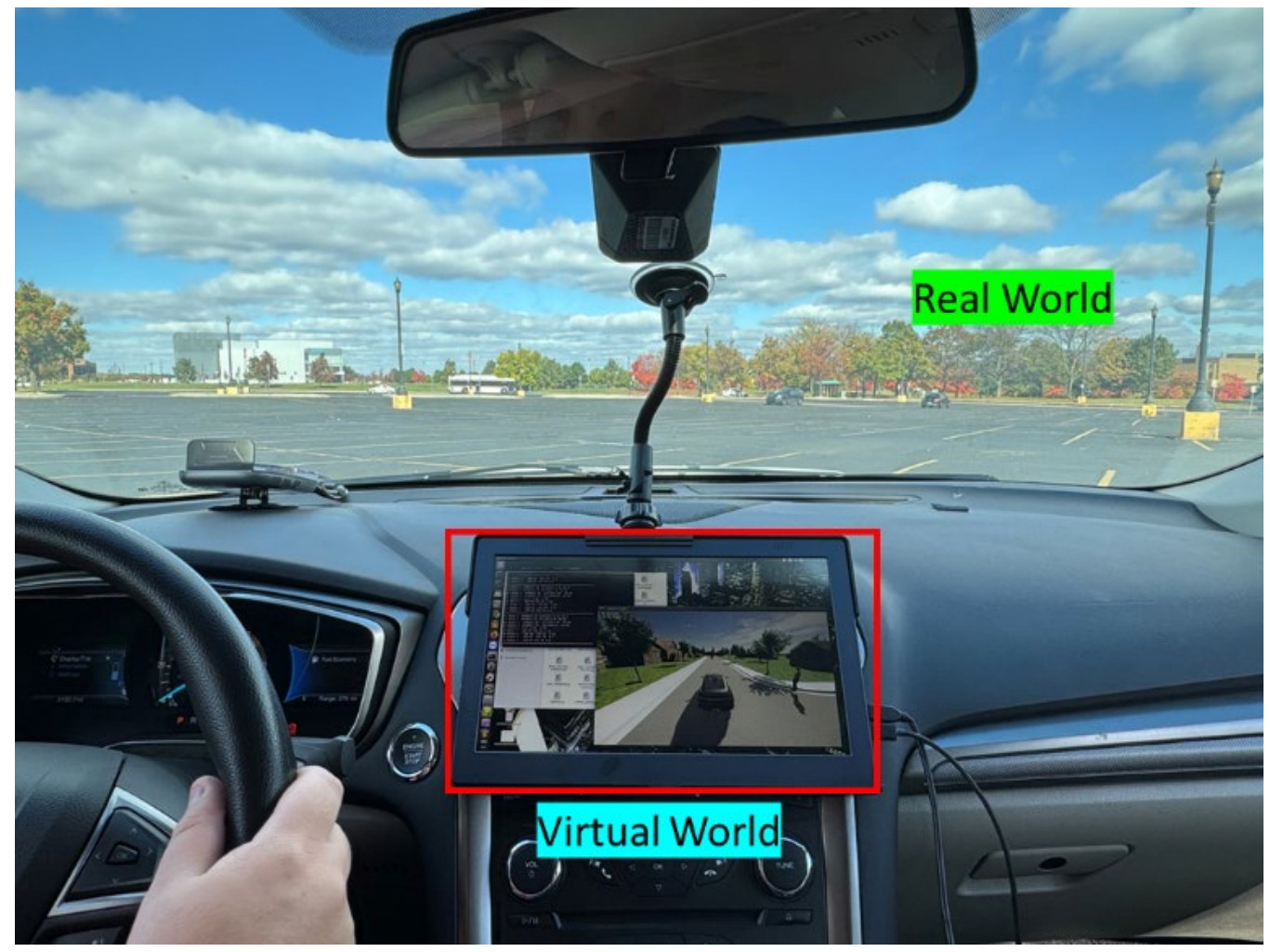

Figure 12. VVE real-virtual motion synchronization

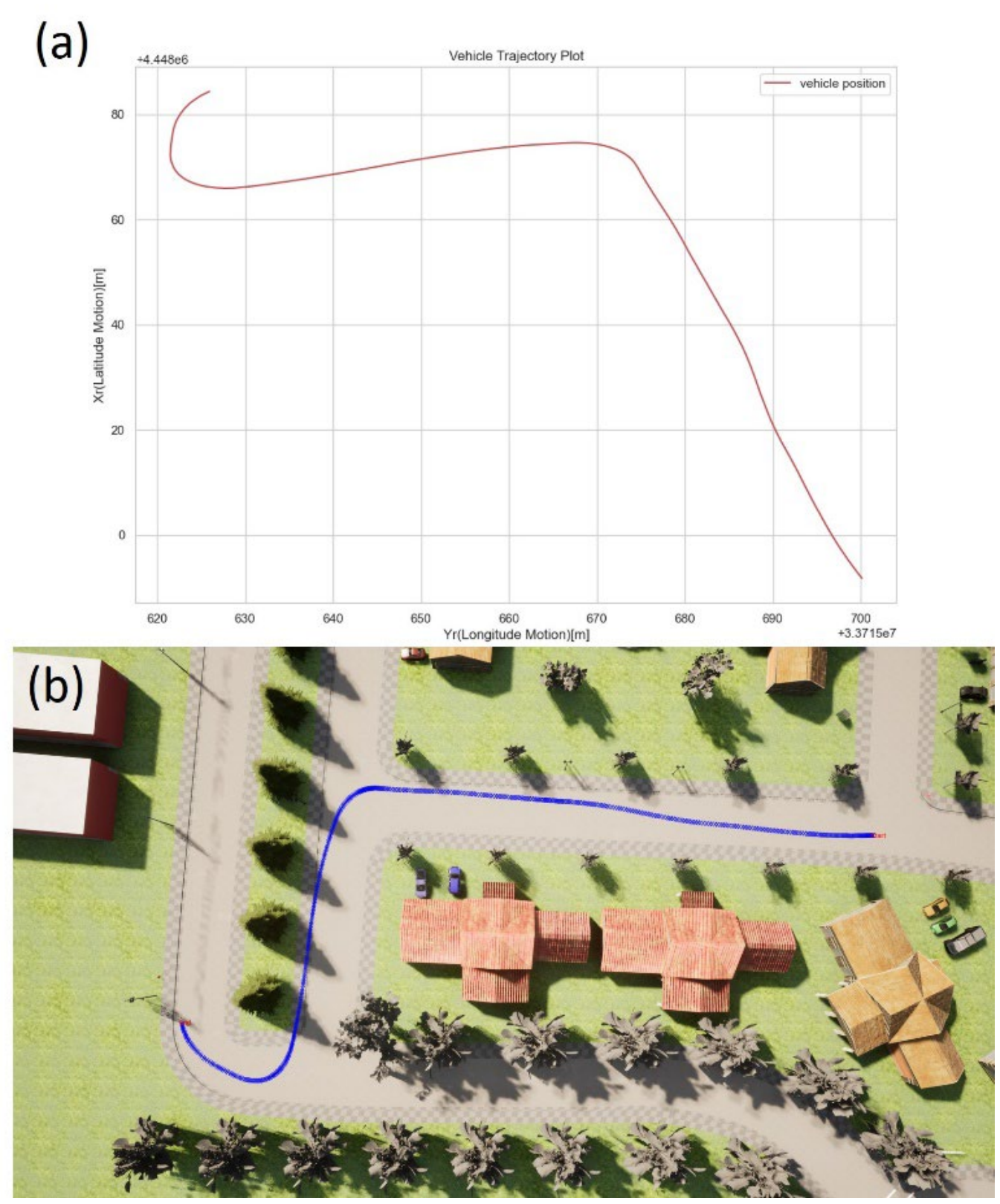

Figure 13. VVE motion synchronization test results: (a) Motion trajectory in the real world; (b) Motion trajectory in the virtual world

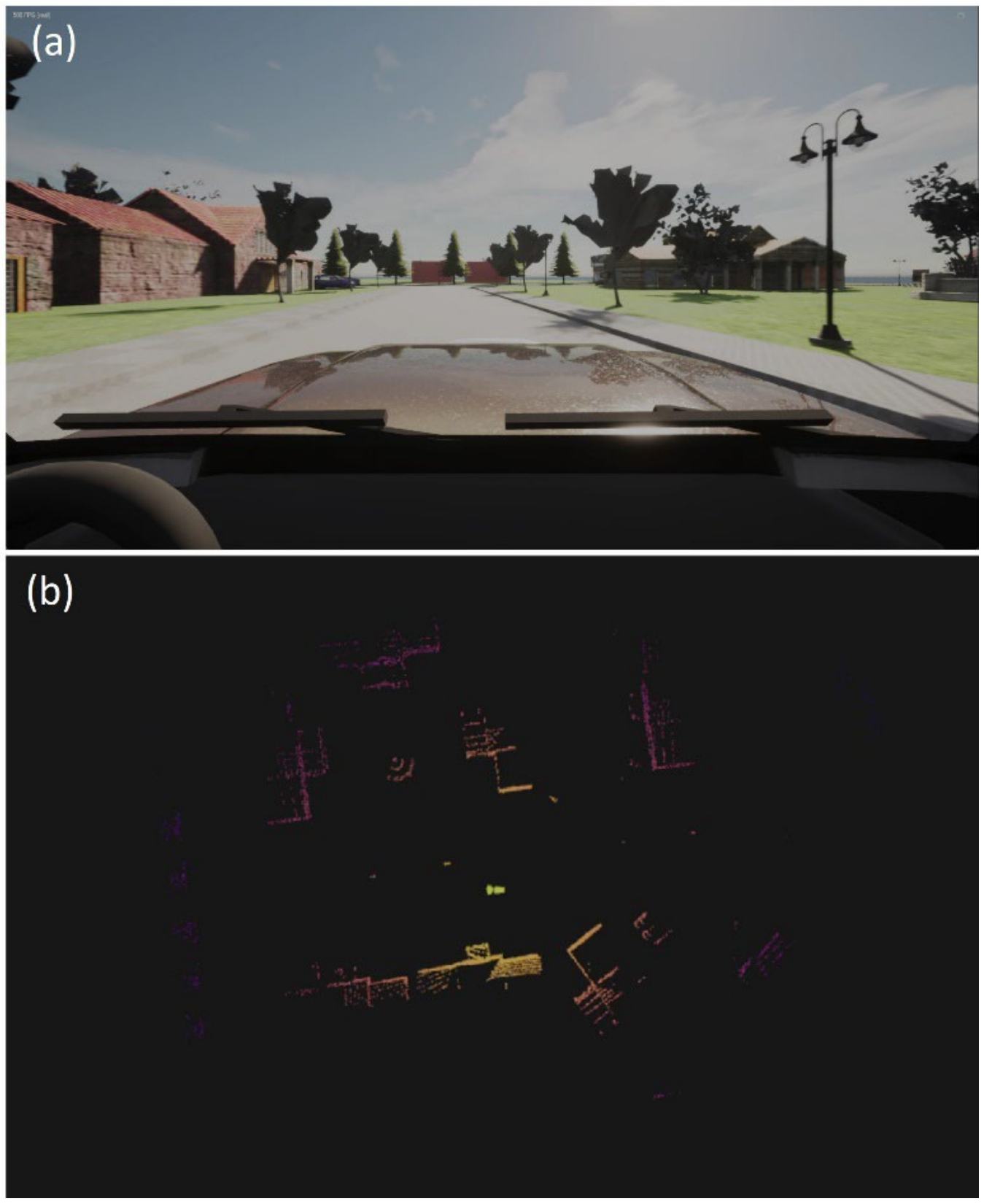

Figure 14. VVE virtual sensor data: (a) RGB camera; (b) LiDAR (figure not to scale)

To demonstrate the Vehicle-to-VRU communication capability with the VVE method, the test is conducted with the setup displayed in Figure 15. The virtual pedestrian is spawned in the virtual environment and the initial position and orientation is set in the virtual environment in relation to the pose of the virtual vehicle. The real pedestrian walks in the vicinity of the real test vehicle while holding the mobile phone with the V2P safety application enabled. The position and heading of the real pedestrian are sent to the virtual environment via Bluetooth connection and after frame transformation/conversion operations, the location and heading of the virtual pedestrian are updated accordingly. It should be noted that the virtual pedestrian location and heading can be easily reset in the virtual environment, so that the real pedestrian can operate at a safe distance from the test vehicle and in a direction that does not cross path with the test vehicle. Such capability is illustrated in Figure 16, where the real pedestrian is walking behind the test vehicle, while the virtual pedestrian is walking in front of the virtual vehicle. It should be additionally noted that during this test, the real vehicle remained stationary as we only aim to compactly illustrate the functionality of the implemented V2P communication by letting the real pedestrian walk very close to the real vehicle.

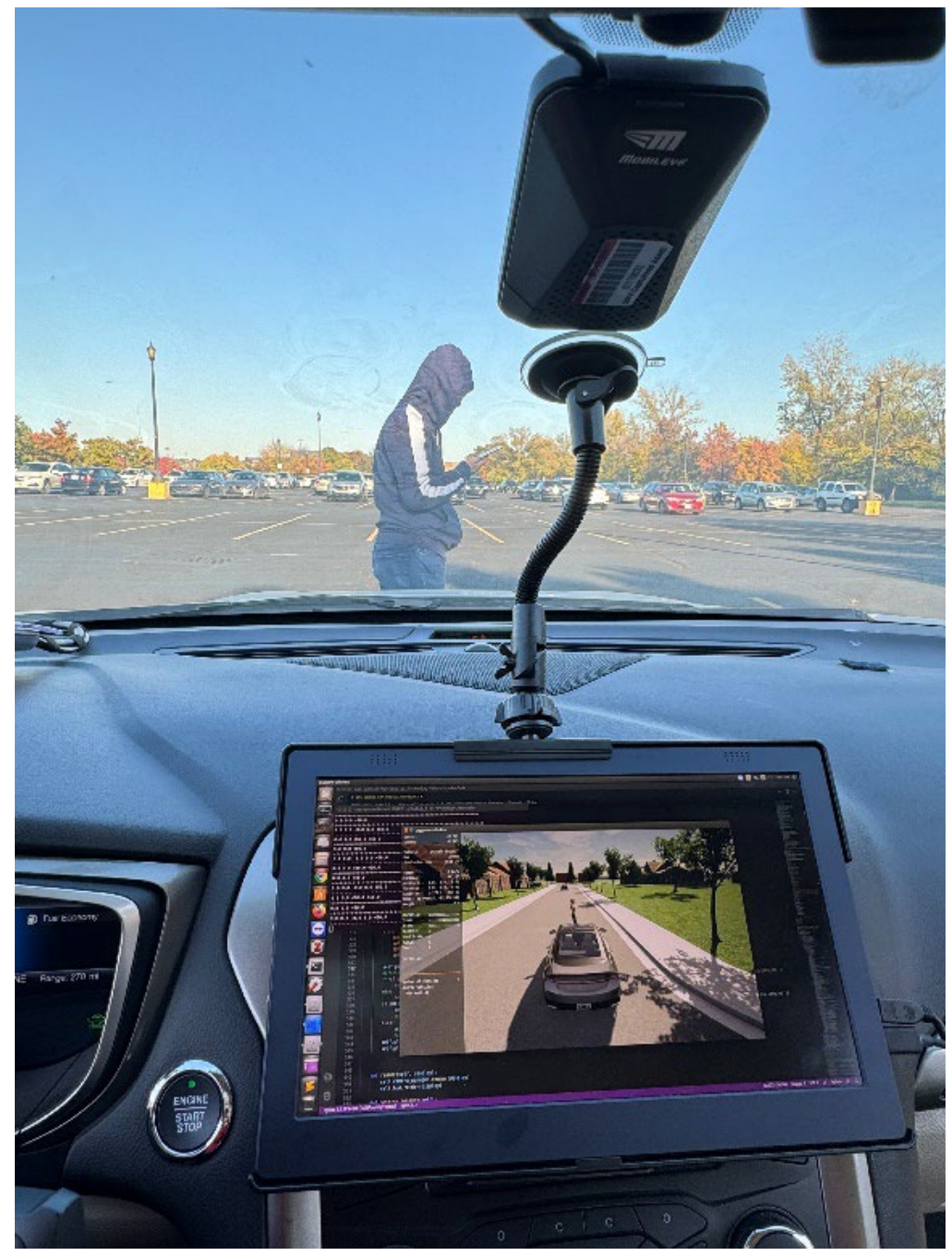

Figure 15. V2P connectivity in VVE test setup

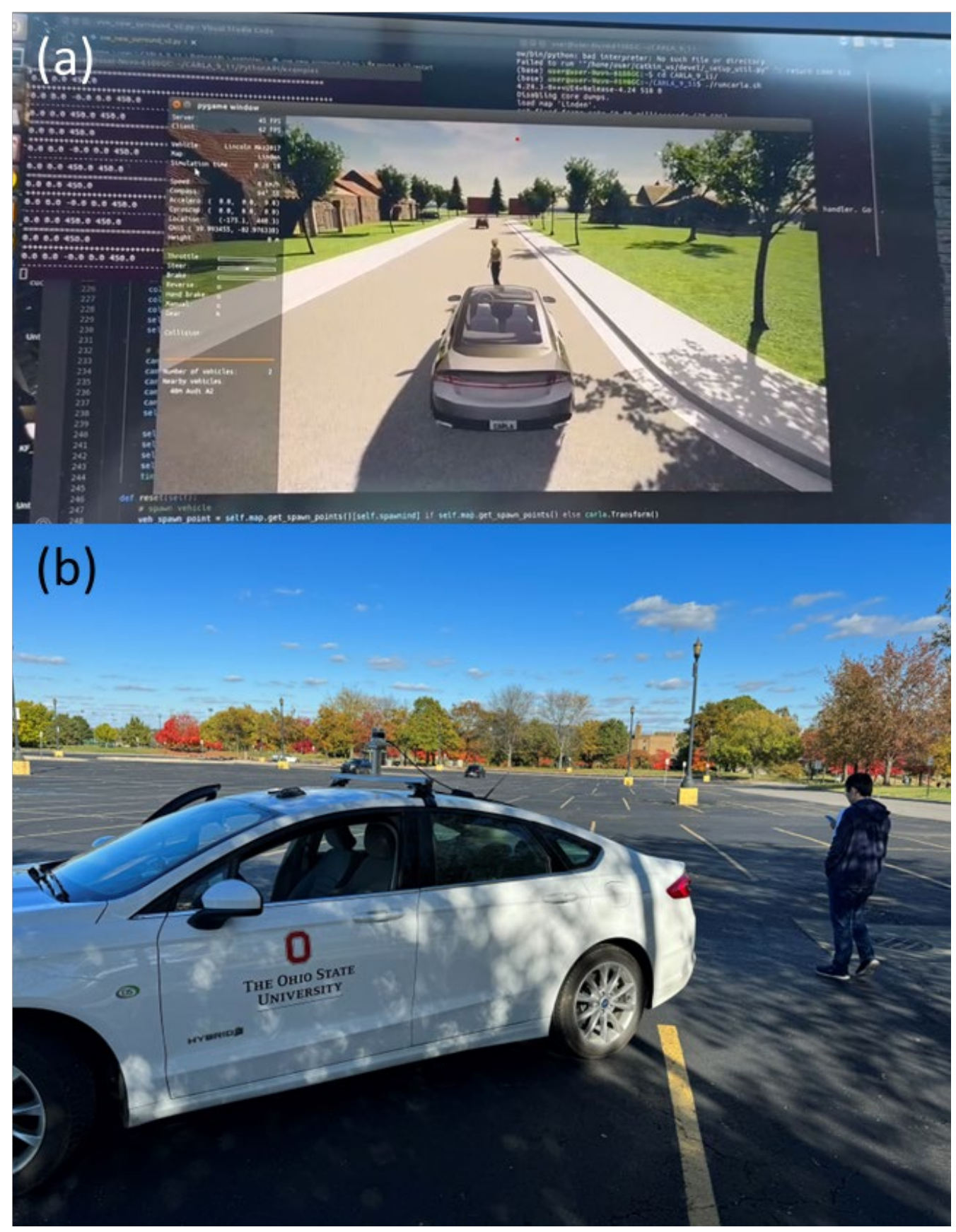

Figure 16. Safe operation of V2P connectivity test: (a) Virtual pedestrian walks in front of the virtual vehicle; (b) Real pedestrian walks behind the real test vehicle

## Summary and Future Work

This paper contributed to a new method for safe, efficient and cost-effective development, demonstration and evaluation of ADAS and CAD systems called Vehicle-in-Virtual –Environment (VVE). This method aims to synchronize the motions of the real test vehicle operating in a safe environment in the real world and those of a virtual vehicle that is embedded in a realistic 3D virtual environment, where different traffic scenarios can be created easily for testing purposes. Implementation details of this method, particularly the frame transformation/conversion operations that guarantees satisfactory motion synchronization are described. A Vehicle-to-VRU communication setup is also implemented to demonstrate the multi-actor capability of this VVE method.

Several future work directions will be explored. One of them is collision avoidance system [19], [20], [21], [22] development using the VVE approach. A more detailed analysis of effects of communication latency and computation delay will also be performed. Additionally, the authors are looking to pursue future development involving extended reality (XR) goggles to immerse the real pedestrian (and possibly the driver) in the virtual environment, the goal of which is to enhance testing realism by allowing the participants to behave more naturally. Preliminary work on displaying the virtual environment in virtual reality VR and augmented reality AR headset has already been conducted. Furthermore, other types of road users such as vehicular road users will be explored to interact with the ego vehicle in the virtual environment. It is possible for other real vehicles to take place in the VVE experiment by sharing the virtual environment. For the ego real vehicle, the other real vehicle needs to be displayed accurately as a virtual vehicle with the correct length, width and shape such that perception sensor data is accurate. Finally, more diverse virtual environments and scenarios (see references [23], [24] for examples) will be generated to evaluate the long-term reliability of the functionalities developed with the proposed VVE method.

## Contact Information

Xincheng Cao: cao.1375@osu.edu

Haochong Chen: chen.9286@osu.edu

## Acknowledgments

This project is funded in part by Carnegie Mellon University's Safety21 National University Transportation Center, which is sponsored by the US Department of Transportation. The authors thank NVIDIA for its GPU donations. The authors thank the Automated Driving Lab at the Ohio State University for its support.